\documentclass[a4paper,conference]{IEEEtran}
\ifCLASSINFOpdf
\else
\fi
\usepackage{times}  
\usepackage{helvet}  
\usepackage{courier}  
\usepackage{url}  
\usepackage{graphicx}  
\usepackage{booktabs}
\usepackage[numbers,sort&compress]{natbib}
\usepackage{threeparttable}
\usepackage{multirow}
\begin{document}
%
\title{Rethinking ReLU to Train Better CNNs}


\author{Gangming Zhao$^{1, 4}$, Zhaoxiang Zhang$^{1, 2, 3, 4*}$\thanks{Corresponding author. (zhaoxiang.zhang@ia.ac.cn)}, He Guan$^{1, 2, 4}$, Peng Tang$^{6}$, Jingdong Wang$^{5}$\\
$^1$Research Center for Brain-inspired Intelligence, CASIA\\
$^2$National Laboratory of Pattern Recognition, CASIA\\
$^3$CAS Center for Excellence in Brain Science and Intelligence Technology\\
$^4$University of Chinese Academy of Sciences\\
$^5$Microsoft Research\\
$^6$School of EIC, Huazhong University of Science and Technology\\
}


%


\maketitle

\begin{abstract}
 Most of convolutional neural networks share the same characteristic: each convolutional layer is followed by a nonlinear activation
 layer where Rectified Linear Unit (ReLU) is the most widely used. In this paper,
 we argue that the designed structure with the equal ratio between these two layers may
 not be the best choice since it could result in the poor generalization ability.
 Thus, we try to investigate a more suitable method on using ReLU to explore the better network architectures.
 Specifically, we propose a proportional module to keep the ratio between convolution and ReLU amount to be N:M (N$>$M).
 The proportional module can be applied in almost all networks with no extra computational cost to improve the performance.
 Comprehensive experimental results indicate that the proposed method achieves better performance on different benchmarks
 with different network architectures, thus verify the superiority of our work.
\end{abstract}


%
\IEEEpeerreviewmaketitle

\section{Introduction}
\label{sec:intro}
Nowadays, with the help of large scale image datasets, \emph{e.g.,} ImageNet \cite{Ref:Russakovsky2015}) and high computing ability of GPU,
deep Convolutional Neural Networks (CNNs) \cite{Ref:Lecun1998} have been dominant in many computer vision applications \cite{zhu2017feature},
especially in image classification \cite{Ref:Krizhevsky2012,Ref:Chatfield14,Ref:Szegedy2015,Ref:Simonyan2015,Ref:He2016,Ref:Zhao2016,Ref:He2016identity}.

One revolutionary contribution of deep learning system is to utilize Rectified Linear Unit (ReLU) \cite{glorot2011deep}
to replace the previous sigmoid and tanh function. Saturated activation functions like sigmoid and tanh lead to the ``vanishing gradient'',
while ReLU prunes the negative values to zero, and retains the positive part that alleviates the vanishing gradient problem.
Moreover, Glorot \emph{et al.} \cite{glorot2011deep} holds the view that ReLU boost the sparsity of the output feature which benefits the performance.
\begin{figure}[t]
\center
	        \begin{minipage}[t]{0.5\linewidth}
	        \includegraphics[width=1\textwidth]{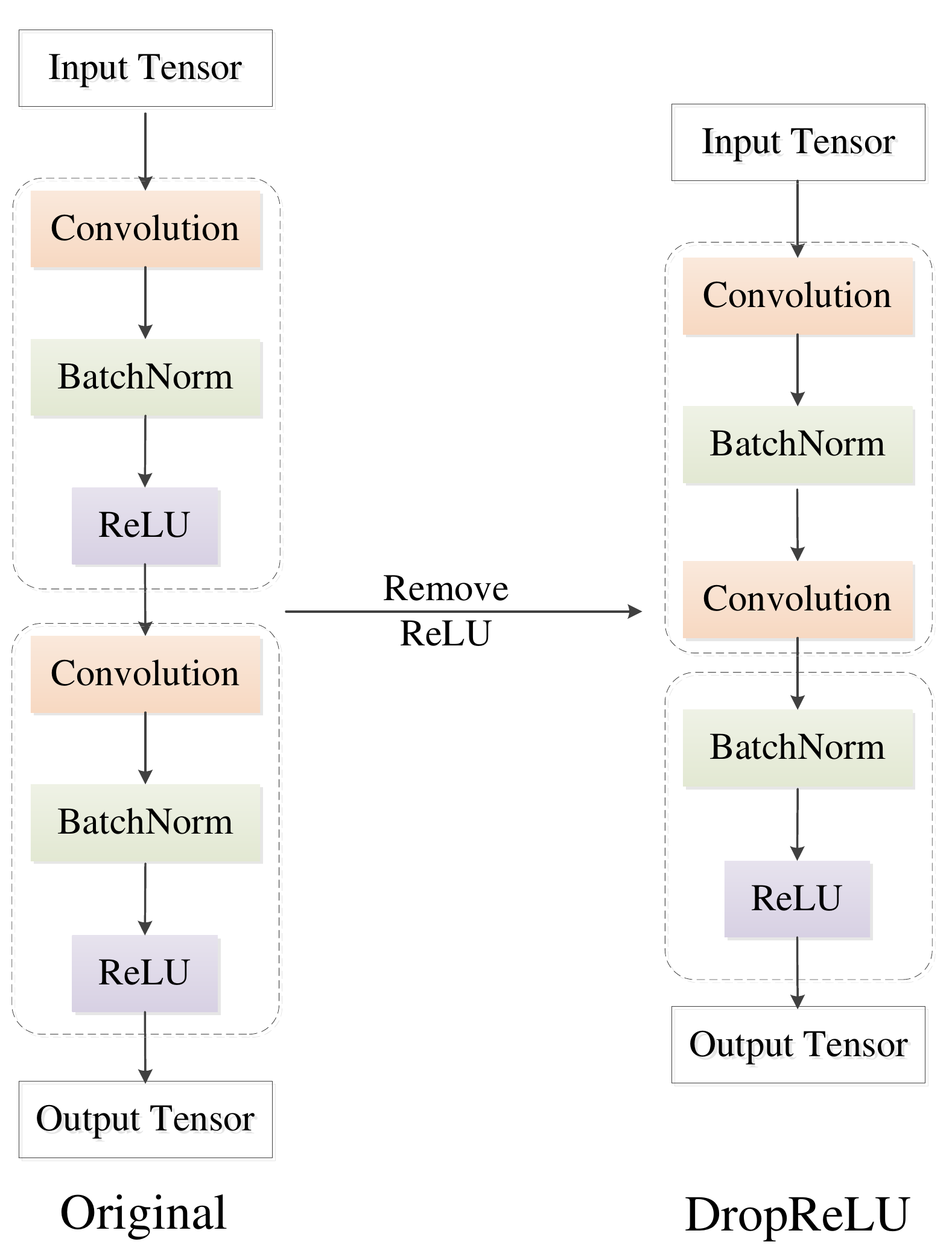}
	        \end{minipage}
\caption{The left Figure shows a part of plain networks,
which is composed of the conventional paired module, the right Figure shows the proportional module, which removes the first ReLU layer.}
\label{VGG}
\end{figure}
Recently,  to achieve better performance, most works focus on network architectures, such as deeper \cite{Ref:Szegedy2015,Ref:Simonyan2015,Ref:He2016} or wider \cite{Ref:Zagoruyko2016,Ref:Zeiler2014} networks. Although there are different architectures, all these networks have the same characteristic: each convolutional layer is followed by an activation function, mostly ReLU. In the previous research, we empirically find that removing some ReLUs in deep CNNs can further improve performance. Thus, we argue that the paired module with the 1:1 ratio between convolution and ReLU may not be the best choice to design network architectures. In this paper, we first propose a proportional module to investigate the more suitable convolution and ReLU ratio, which can assist researchers to design the better network architectures. We call the module with equal ratio between convolution and ReLU module as paired module. Correspondingly, we call the N:M ratio between convolution and ReLU module as proportional module. Paired module is only composed of the non-linear unit. But the linear unit can help different networks expressing the distribution of data better since it use information effectively. For example, the feature tensors after the linear unit can recover the information discarded by ReLUs, which is benefit to the whole network. As shown in Figure \ref{VGG}, our approach removes the first ReLU layer of the basic module. We call this modified module as 2-1 proportional module.

The contribution of this paper includes three parts:
\begin{itemize}
  \item We first propose the proportional module and apply it to CNN models.
  \item Numerous experimental results verify the reasonableness of our assumption. The proportional module can further improve the performance of the specific network architectures.
  \item The proportional module can be applied in almost all networks with no extra computational cost. This simple method can help researchers to design the better network architectures.
\end{itemize}

\section{Related Work}
With the rapid development of deep learning, many efforts have been made to improve the learning algorithms for CNN, such as ReLU for improving feature representation of model. Dropout \cite{Srivastava2014Dropout} and Batch Normalization \cite{ioffe2015batch} are useful for both performance and efficiency improvement. In addition,
GoogLeNet \cite{szegedy2015going}, ResNet \cite{he2016deep} \cite{Ref:He2016identity}, Deep Fusion Network (DFN) \cite{Ref:Zhao2016}, Xception \cite{chollet2016xception}, ResNeXt \cite{xie2016aggregated} and IGCNets \cite{zhanginterleaved} have fully verified that it is valuable to effectively train a deeper or wider network with better performance. Furthermore, many researches like \cite{Ref:stewart2017label,Ref:bacon2017option} explore the inner mechanism and more fundamental theory of deep learning, which effectively improve the interpretability of deep learning.

\subsection{Rectifier Units}
Among these advancements, ReLU is one of several factors to the success of deep learning. Despite the great performance improvement, there have still been many recent improvements of activation functions containing leaky rectified linear (LReLU) \cite{maas2013rectifier}, randomized leaky rectified linear (RReLU) \cite{maas2013rectifier} and parametric rectified linear (PReLU) \cite{he2015delving}, which are useful to the optimization of network so as to improve the performance. In addition, Shang \emph{et al.} \cite{shang2016understanding} proposed concatenated rectified linear (CReLU) to alleviate the problem of negative correction in convolutional layers. Besides, Zhu et al. [26] introduced a compositional activation function for DCNN. Furthermore, exponential linear units (ELUs) \cite{clevert2015fast} were proposed to speed up learning in deep neural networks and acquire higher classification accuracies, but ELUs increased the computational complexity of networks. In contrast with ELUs, our proposed method can effectively improve the performance of various CNN structures with no extra computational cost.
\subsection{Batch Normalization}
Batch Normalization (BN) was proposed by Sergey \emph{et al.} \cite{ioffe2015batch} to alleviate the internal covariate shift problem. As a regularizer, Dropout is gradually replaced by BN. Currently, BN is widely applied in almost all CNN models.

BN allows researchers to create deeper and wider networks, and it is one of important keys to the great success of ResNet. Generally, networks are composed of the sequence of CONV-BN-ReLU. Also, the sequence of BN-ReLU-CONV can further improve the performance of some networks like ResNet. Proportional module also can be considered as a regularization method to alleviate the overfitting of models. It explores more flexibly combinational methods of CONV, BN and ReLU. For example, the sequence of CONV-BN-CONV-BN-ReLU is a 2:1 proportional module, it effectively improves the performance of networks in our experiments. But whether is the combinational method the best module in CNN models. We need to do more experiments to explore the influence of different proportional module on networks.

In addition, BN have great influence on proportional module. For example, if we remove the first ReLU and BN in
building block of ResNet \cite{he2016deep},
the two convolutional layers become a convolution with larger receptive field. Similarly, if removing the last ReLU and BN,
pre-activation building block is also the same.
The two convolutional layers do not become one convolution because the existence of the first BN. For another case, if we remove the first ReLU and BN of pre-activation building block
or the last ReLU and BN of building block, the phenomenon of the first case will not appear, whether or not is the first BN of pre-activation building block or the last BN of building
block redundant. We still need to conduct more experiments and theoretical analysis to explore the influence of this BN.

\section{OUR APPROACH}
\begin{figure}[t]
\flushright
	        \begin{minipage}[t]{1\linewidth}
	        \includegraphics[width=1\textwidth]{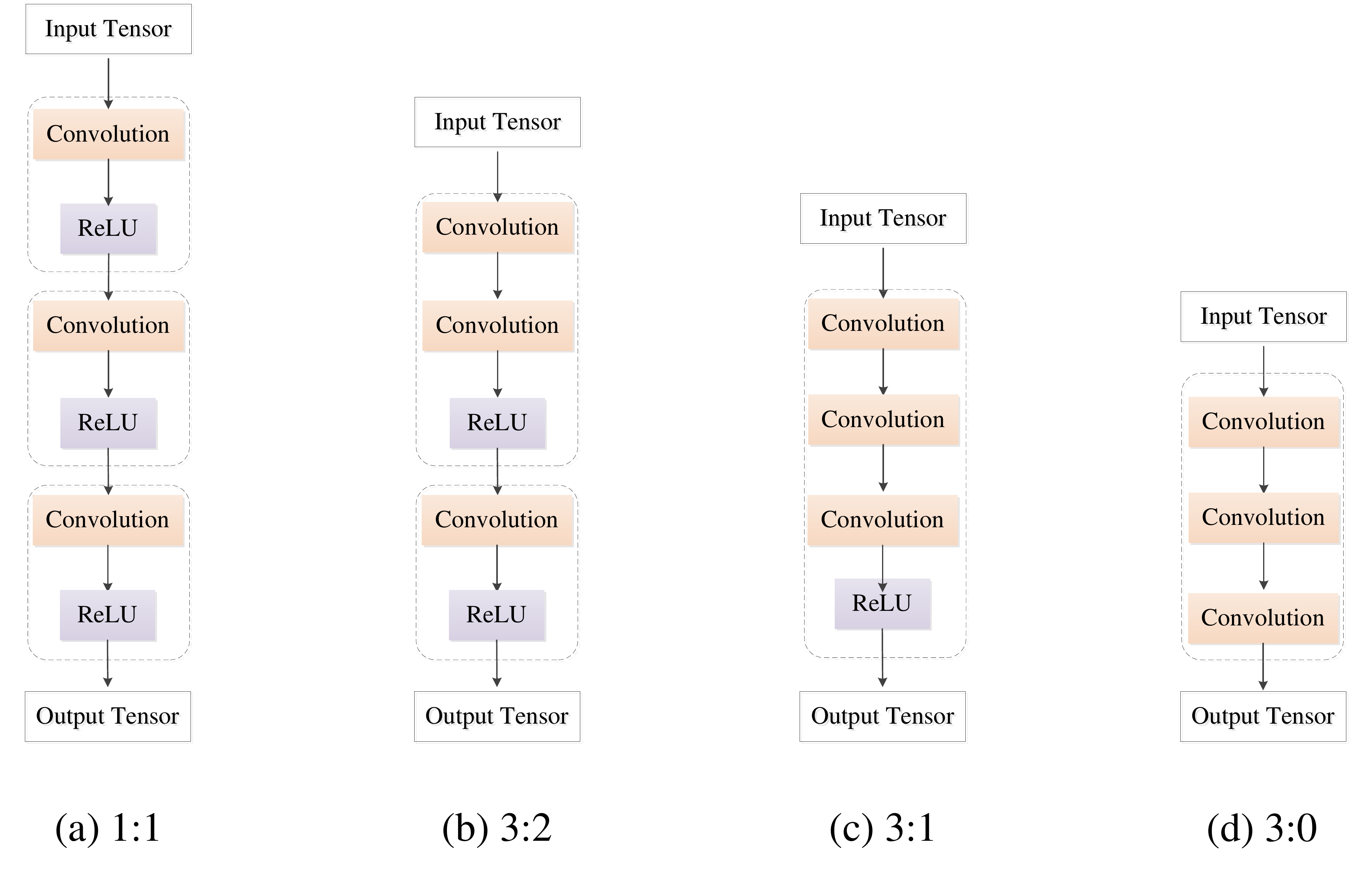}
	        \label{fig:notification-sys:a}
	        \end{minipage}
\caption{(a): The paired module. (b), (c): The different proportional modules. (d): If removing all ReLUs, the model becomes a linear module, it is an extreme example and it can not work because of without non-linear function.}
\label{Ensemble}
\end{figure}
\subsection{Proportional Module}
The current popular networks consist of multiple stacked paired modules. Paired module keeps the ratio between convolution and ReLU amount to be 1:1.
We find that the conventional module could result in network overfitting and poor generalization ability.
We empirically find that reducing the number of ReLUs can be helpful for ameliorating the overfitting in CNN.
In our paper, we adopt a fire-new strategy to regularize CNN by utilizing the proposed proportional module.
Proportional module integrates the linear and non-linear information into models to form richer representations of data.
Furthermore, He \emph{et al.} \cite{he2016deep} \cite{Ref:He2016identity} proposed ResNet which can be considered as another method to combine two types of information.
We apply proportional module in ResNet, which can further improve the classification performance. In a word, proportional module can not only act as a regularizer
but also as a method which can reduce the information loss in forward and backward prorogation process using ReLU. Figure \ref{Ensemble} shows some different proportional modules and paired module.

\begin{figure}[!htp]
\flushright
	        \begin{minipage}[!htp]{1\linewidth}
	        \includegraphics[width=1\textwidth]{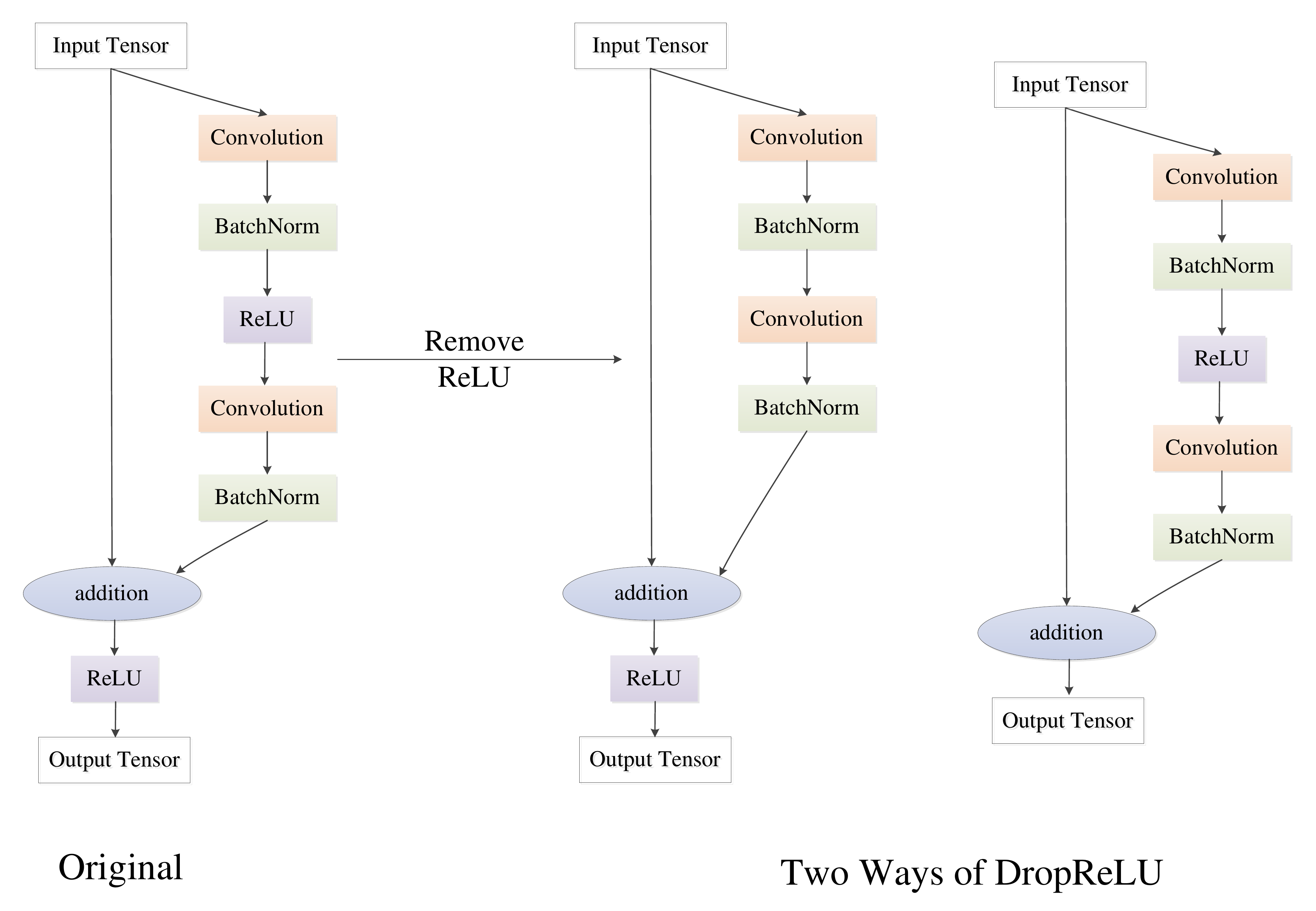}
	        \label{fig:notification-sys:a}
	        \end{minipage}
\caption{The building block of ResNet vs. two ways of applying proportional module in building block.}
\label{ResNet}
\end{figure}
\subsection{Combinatorial Method for Diverse Modules}
\begin{figure}[!htp]
\flushright
	        \begin{minipage}[!htp]{1\linewidth}
	        \includegraphics[width=1\textwidth]{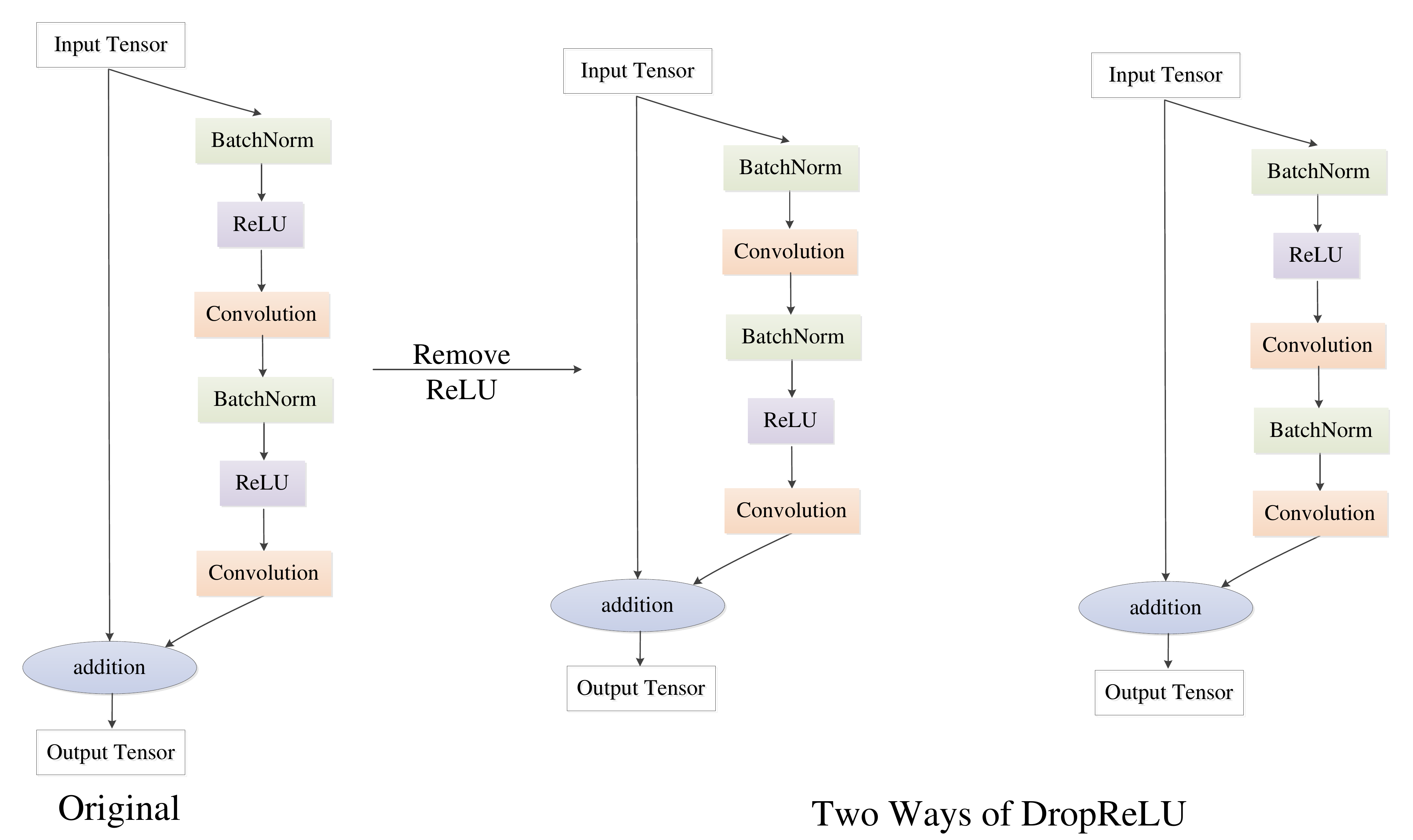}
	        \label{fig:notification-sys:a}
	        \end{minipage}
\caption{The pre-activation building block of ResNet vs. two ways of applying proportional module in the pre-activation building block.}
\label{Pre-ResNet}
\end{figure}

\begin{figure}[!htp]
\flushright
	        \begin{minipage}[!htp]{1\linewidth}
	        \includegraphics[width=1\textwidth]{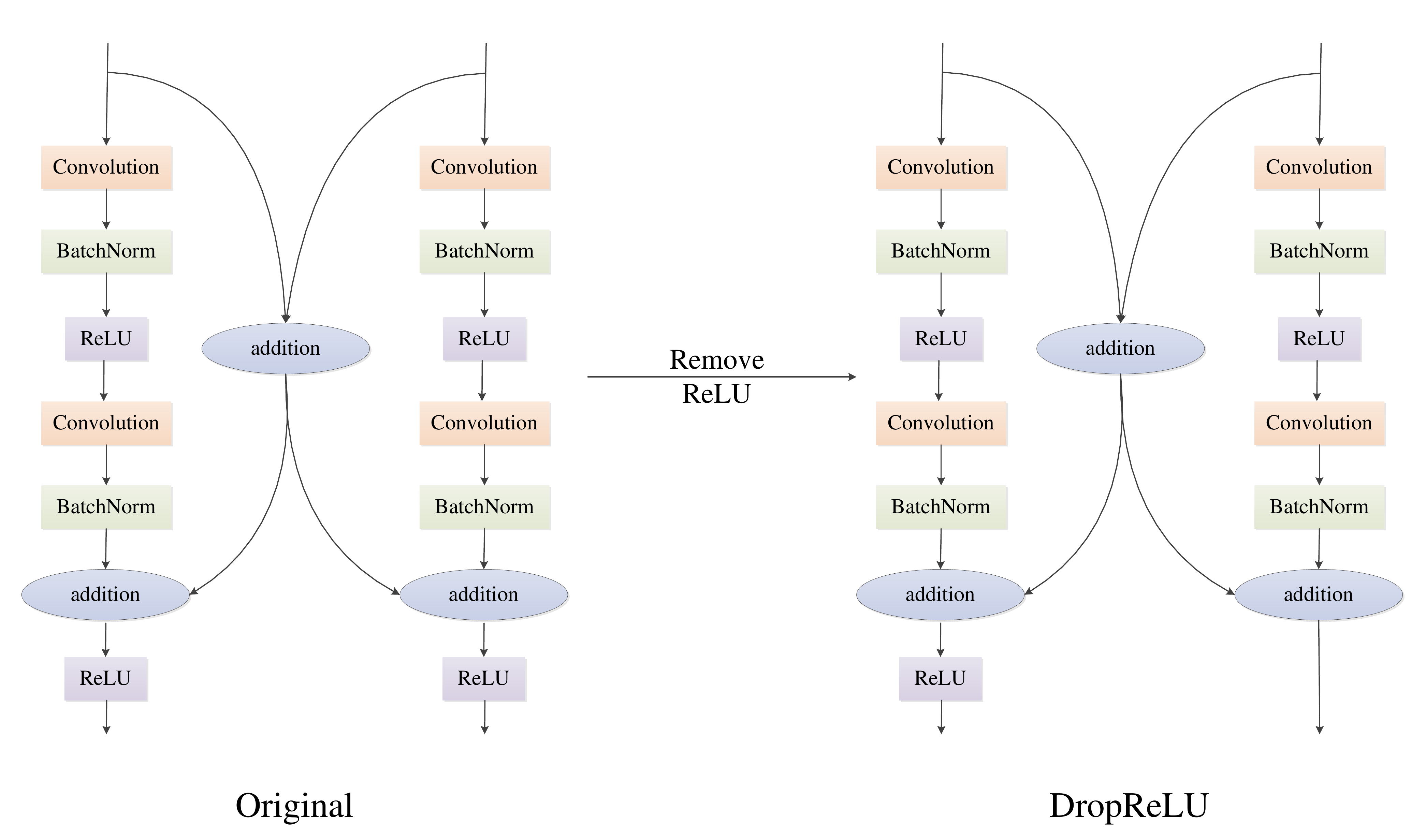}
	        \label{fig:notification-sys:a}
	        \end{minipage}
\caption{The merge and run block of DFN vs. two ways of applying proportional module in the merge and run block .}
\label{DFN}
\end{figure}

In this section, we will introduce the different combinatorial method of diverse modules in networks including plain network in Figure \ref{VGG}, ResNet in Figure \ref{ResNet}, Pre-ResNet in Figure \ref{Pre-ResNet} and Deep Fusion Network in Figure \ref{DFN}.
Paired module has two forms:
(1) a convolutional layer is followed a ReLU that we call this module as Post-Paired Module or Paired Module and
(2) a convolutional layer is placed in the after of ReLU that we call this module as Pre-Paired Module.
Also, there are two identical forms in proportional module.
We apply these four modules in plain network, ResNet and Deep Fusion Network (DFN), respectively.

The building block is the first residual module proposed by He \emph{et al.} \cite{he2016deep}. These blocks are shown in Figure \ref{ResNet} and \ref{Pre-ResNet}. For building block, the method of applying proportional module has two styles that shown in Figure \ref{ResNet}.

Also, for the pre-activation building block, there are two styles of applying proportional module. We use the pre-activation building block to conduct experiments on ResNet because we find that removing ReLU in
 building block has the similar results in pre-activation building block. In experimental section, we only show the results of
removing the first ReLU on pre-activation building block since
we find that removing the first ReLU of pre-activation block achieve the higher performance than removing the second one.
But why this result occurs, we also find the similar results in the pre-activation bottleneck block. The detailed analysis is shown in experimental section.

Deep Fusion Network (DFN) \cite{Ref:Zhao2016} is proposed by Zhao \emph{et al.}, who introduced a merge and
run fusion scheme to combine two sub-networks, which are all called
as deeply merge-and-run fused network (DFN-MR). DFN-MR performs superiorly to existing popular networks like ResNet. In our experiments, we utilize proportional module to replace paired module. There are many methods to replace paired module. We list one replaced method shown in Figure \ref{DFN}.

\section{Experiments}

In this section, we will introduce our experiments on plain network, ResNet \cite{Ref:He2016,Ref:He2016identity} and Deep Fusion Network (DFN) \cite{Ref:Zhao2016}.
In the first section, we will briefly describe the datasets. Then in next section, we explore the influence of our method by applying proportional module in plain network. In addition, we adopt two popular networks including DFN and ResNet to combine paired module and proportional module for further exploring the better performance. All the tables show the average classification accuracy from 5 runs and the standard deviation (mean $\pm$ std).

\subsection{Datasets}
In our experiments, we use two datasets that are commonly used for image
recognition: CIFAR-10 and CIFAR-100. All the two datasets are detailed as follows.

\textbf{CIFAR-10}: The CIFAR-10 dataset has 50,000 images for training and 10,000
for testing. All images are tiny RGB images with a size of 32 $\times$ 32.

\textbf{CIFAR-100}: The CIFAR-100 dataset is similar to the CIFAR-10, except that
it has 100 classes.

\subsection{The Various Networks}
The setting for the different networks are as follows:

We modify the plain network, which is introduced in \cite{he2016deep}, to fit CIFAR-10 and CIFAR-100. The detail about the three networks are introduced in \cite{Ref:Zhao2016}.
Like \cite{Ref:Zhao2016}, we use the SGD algorithm with the Nesterov momentum to train all the models for 400 epochs on CIFAR-10/CIFAR-100, both with a total mini-batch size 64.
The detail of the experimental setting we used is also introduced in \cite{Ref:Zhao2016}.
\begin{table}[t]

	\centering
\caption{Comparison between 1:1 convolution and ReLU ratio and 2:1 convolution and ReLU ratio in plain network on CIFAR-100.}
	\label{comparison}
	\begin{tabular}{|c|c|c|}
		\hline
		Model & Depth & Accuracy \\
        \hline
        {Paired Module}  & {38}  & {64.04 $\pm$ 0.14} \\
        \hline
        {2-1 Proportional Module} & {38} & {64.25 $\pm$ 1.44} \\
        \hline
        {Paired Module}   & {62}  & {56.88 $\pm$ 1.16} \\
        \hline
        {2-1 Proportional Module} & {62}  & {\textbf{60.86 $\pm$  1.18}} \\
        \hline
        {Paired Module}   & {84} &  {44.16 $\pm$ 0.57} \\
        \hline
        {2-1 Proportional Module}  & {84}  & {\textbf{55.98 $\pm$ 2.59}} \\
        \hline
	\end{tabular}
\label{plaint}
\end{table}

\subsubsection{The Plain Network}
In this section, we utilize plain network to verify that the N:M (N$>$M) convolution and ReLU ratio is indeed effective.
We argue that although ReLUs alleviate the problem of vanishing gradient caused by sigmoid and tanh activation functions,
the redundancy of ReLUs is an important reason why the accuracy of deeper network gets saturated. We conduct comparative
experiment of the 2:1 convolution and ReLU ratio and 1:1 ratio to explore the effectiveness of proportional module.
To show the ratio of convolution and ReLU, we call these modules Paired Module and 2:1 Proportional Module.
In addition, 2:1 Proportional Module means we remove the last ReLU of the 1-th, 3-th, 5-th and etc.

Table \ref{plaint} shows that our experimental results. He \emph{et al.} \cite{Ref:He2016} had
verified that the degradation problem becomes more serious with the network depth increasing.
From Table \ref{plaint}, we can find that a proportional module with 2:1 convolution and ReLU ratio can alleviate
degradation problem effectively. And 2:1 proportional module improves the performance of the conventional paired module
by 3.98\% and 11.82\% in 62 and 84-layer plain networks.
\begin{table}[b]

	\centering
	\caption{Comparison between 1:1 and 2:1 convolution and ReLU ratio in pre-activation block on CIFAR-10/100.}
	\label{comparison}
	\begin{tabular}{|c|c|c|}
		\hline
		Depth / C-R & 1:1 Pre-Paired& 2:1 Pre-Proportional\\
        \hline
        \multicolumn{3}{|c|}{CIFAR-10} \\
        \hline
       {62}  & {93.59 $\pm$ 0.20} & {\textbf{93.97 $\pm$ 0.23}} \\
        \hline
        \multicolumn{3}{|c|}{CIFAR-100} \\
        \hline
        {62}  & {71.43 $\pm$ 0.32} & {\textbf{72.19 $\pm$ 0.18}}  \\
        {110} & {72.46 $\pm$ 0.10} & {\textbf{73.41 $\pm$ 0.10}} \\
        {164} & {73.00 $\pm$ 0.31} & {\textbf{74.12 $\pm$ 0.22}}  \\
        \hline
	\end{tabular}
\label{pre-build}
\end{table}

\begin{figure}[b]
\centering
	        \begin{minipage}[t]{1\linewidth}
	        \includegraphics[width=1\textwidth]{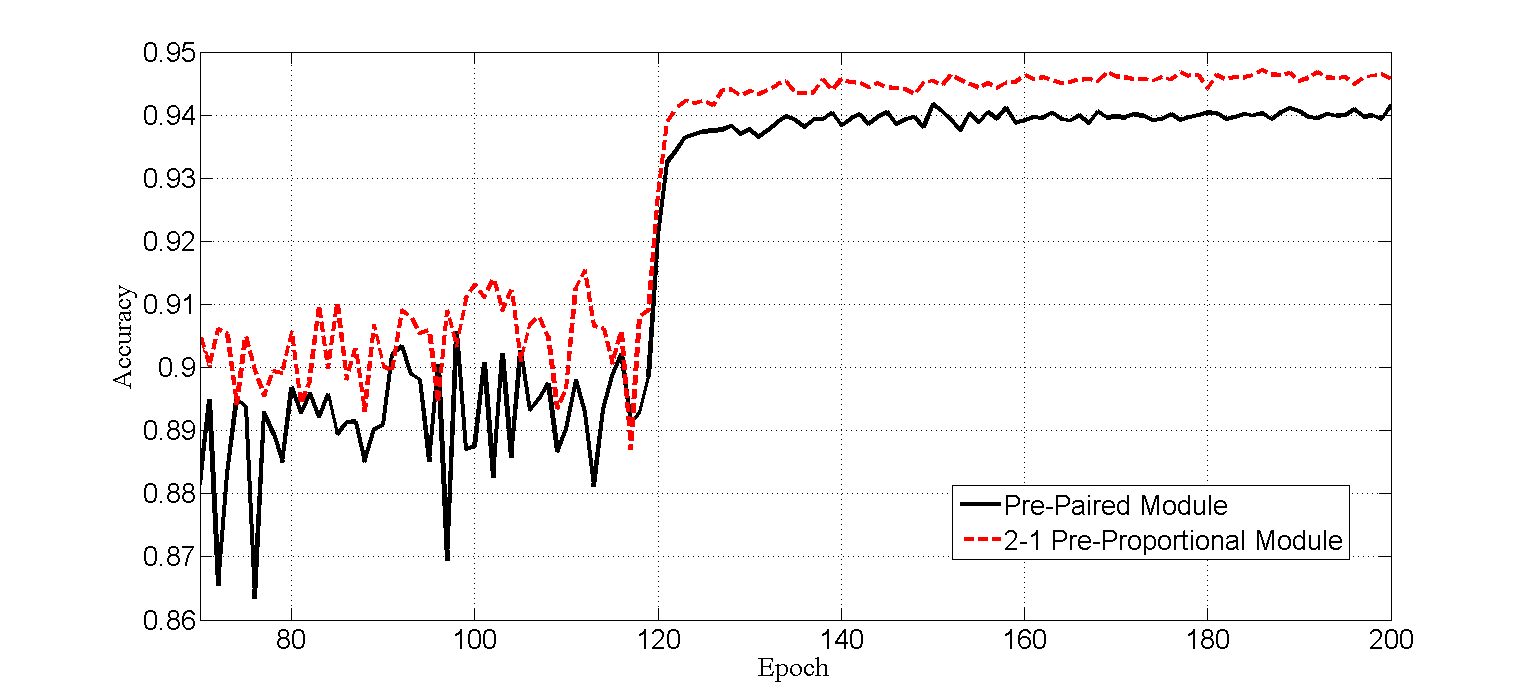}
	        \label{fig:notification-sys:a}
	        \end{minipage}
            \caption{The accuracy of 110-layer ResNet with pre-paired module and pre-proportional module on CIFAR-10. Pre-proportional module means we remove the first ReLU in pre-activation building block.}
            \label{110ResNetCIFAR}
\end{figure}
\subsubsection{ResNet}
In this section, we use building block and pre-activation bottleneck block for exploring the influence of different convolution and ReLU ratio.
Specifically, we adopt the various combinational method of modules to investigate whether N:M (N$>$M) convolution and ReLU module can help improving
the generalization of networks. We have introduced these models in section \emph{Combinatorial Method for Diverse Modules}.
In this section, we will introduce the experimental results of these modules in different residual modules.

\textbf{Building Block}: Table \ref{pre-build} shows the results of pre-activation building block of
ResNet on CIFAR-10 and CIFAR-100. Among of Table \ref{pre-build},
C-R means the convolution and ReLU ratio.
1:1 Pre-Paired Module and 2:1 Pre-Proportional Module mean two different ratios in pre-activation building block.
In addition, for 2-1 Pre-Proportional Module, we remove the first ReLU in pre-activation block.
For 62-layer ResNet, 2:1 Pre-Proportional Module achieves 93.97 recognition rate that is higher than the corresponding baseline on CIFAR-10.
For CIFAR-100, this strategy also achieves the better performance in 62-layer ResNet.
Then we set the depth of network as 110 and 164, respectively.
We find that our method improves the recognition rate by 0.95 and 1.12 on CIFAR-100.
In addition, Figure \ref{110ResNetCIFAR} shows the test recognition rates of pre-activation building block in 110-layer ResNet with 1:1 and 2:1 convolution and ReLU ratio as a
function of number of epoches on CIFAR-10. As shown in Figure \ref{110ResNetCIFAR}, we can find that 2:1 Pre-Proportional Module can further improve the performance of traditional 110-layer ResNet.

\begin{figure}[t]
\centering
	        \begin{minipage}[t]{1\linewidth}
	        \includegraphics[width=1\textwidth]{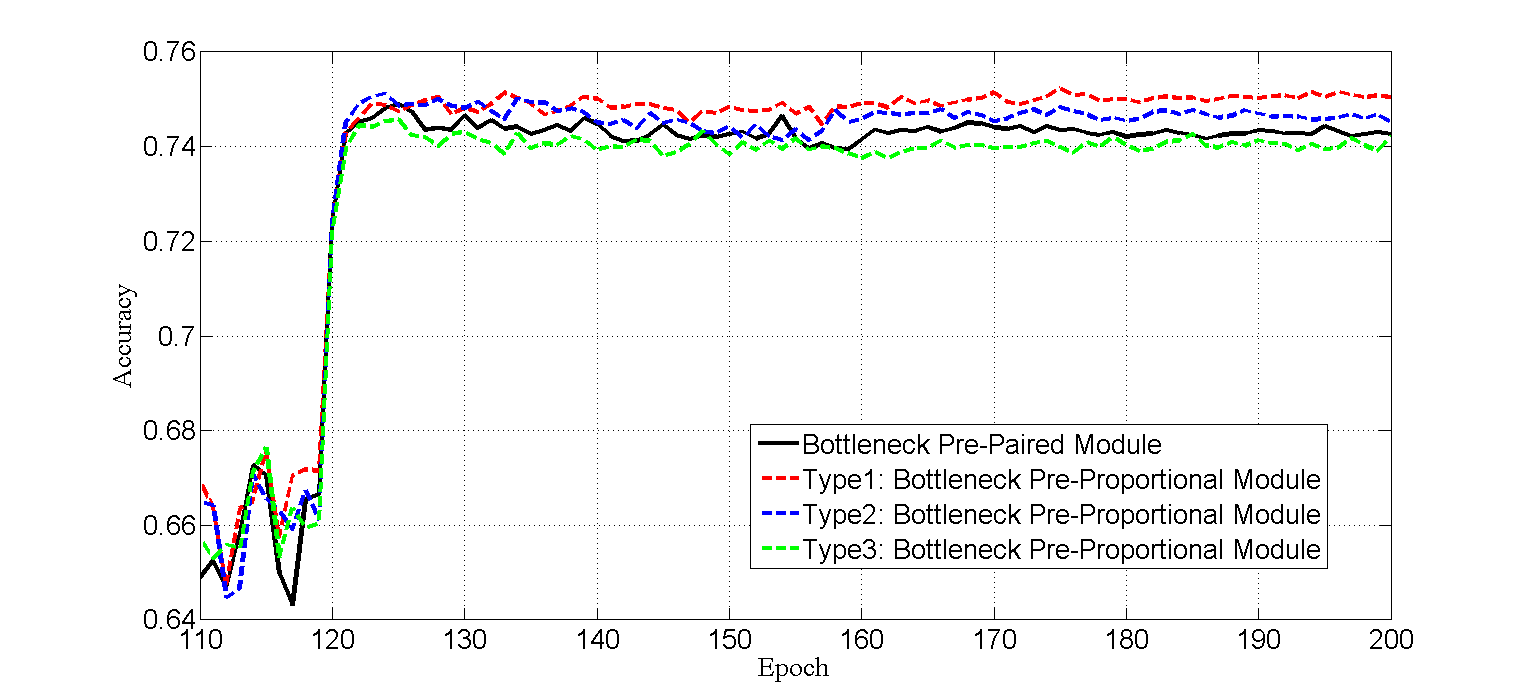}
	        \label{fig:notification-sys:a}
	        \end{minipage}
\caption{The accuracy of 110-layer ResNet with bottleneck pre-paired nodule (baseline) and bottleneck pre-proportional module on CIFAR-100. Type 1, 2
, and 3 denote Removing the first, second, and third ReLU in pre-activation bottleneck block, respectively.}
\label{110-ResNet-CIFAR100}
\end{figure}

\textbf{Pre-activation Bottleneck Block}: Figure \ref{110-ResNet-CIFAR100} shows the test recognition rates of pre-activation bottleneck block in 110-layer ResNet
with 1:1 and 3:2 convolution and ReLU ratio as a function of number of epoches on CIFAR-100.
Among of Figure \ref{110-ResNet-CIFAR100}, Bottleneck Pre-Paired Module means the conventional pre-activation bottleneck block and
Bottleneck Pre-Proportional Module means we remove one of three ReLUs.
So, it has three types, Type 1, 2, 3 correspond to remove the first, second and third ReLUs, respectively.
As shown in Figure \ref{110-ResNet-CIFAR100}, we can find that Type 1, 2 of Bottleneck Pre-Proportional Module improve the performance of baseline and
experimental results show that removing the first ReLU yields the best accuracy.
But the performance of Type 3 Bottleneck Pre-Proportional Module is reduced.

 \begin{figure}[!htp]
\centering
	        \begin{minipage}[t]{1\linewidth}
	        \includegraphics[width=1\textwidth]{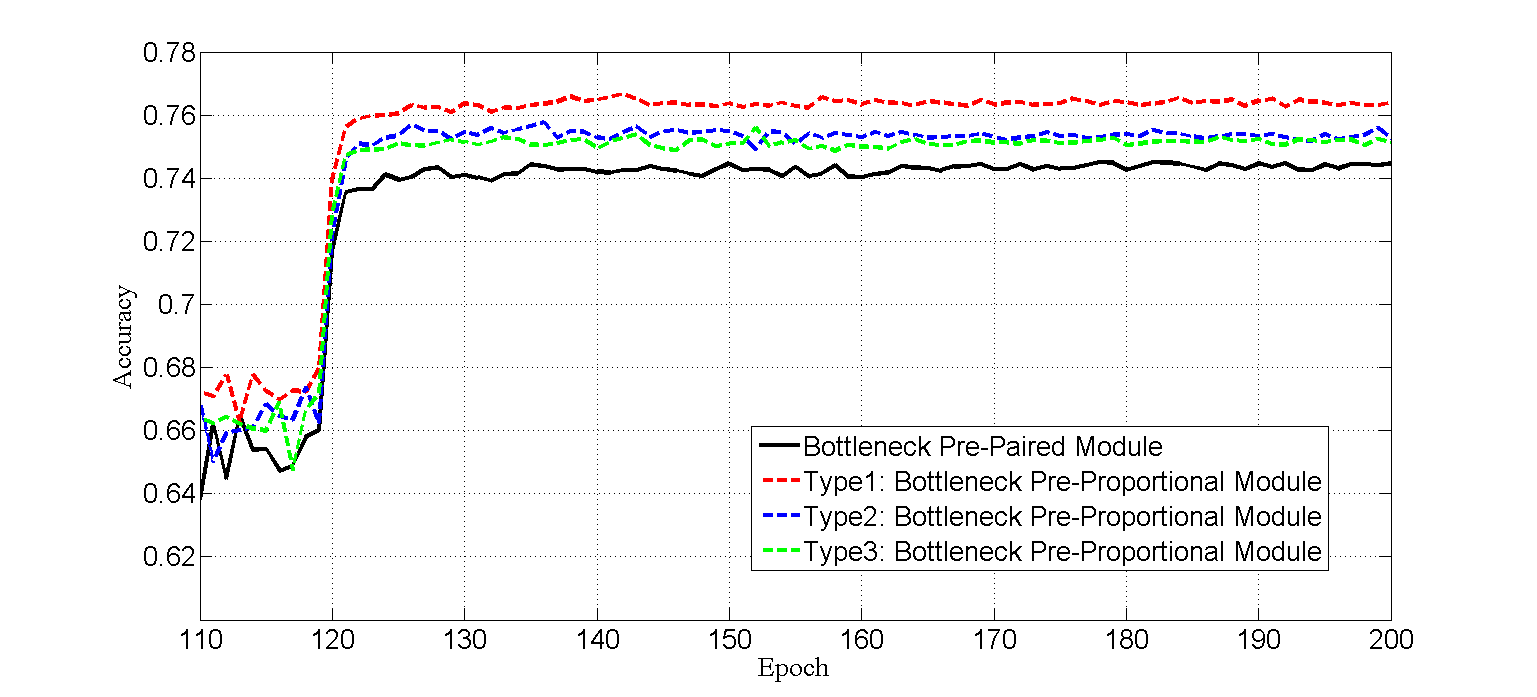}
	        \label{fig:notification-sys:a}
	        \end{minipage}
\caption{The accuracy of 164-layer ResNet with bottleneck pre-paired module (baseline) and bottleneck pre-proportional module on CIFAR-100. Type 1, 2
, and 3 denote Removing the first, second, and third ReLU in pre-activation bottleneck block, respectively.}
\label{164-ResNet-CIFAR100}
\end{figure}

In addition, as shown in Figure \ref{164-ResNet-CIFAR100}, we can find that
although the experiments in 164-layer ResNet have the similar results compared with 110-layer ResNet,
removing the third ReLU also improves the accuracy of baseline.
Furthermore, we find the improvements of 164-layer ResNet with 3:2 convolution and ReLU ratio are higher than the improvements of 110-layer ResNet.

\subsubsection{Deep Fusion Network}
In this section, we utilize the merge and run block to explore the
influence of different convolution and ReLU ratio on Deep Fusion Network.
The module we used is shown in Figure \ref{DFN}.
Among of it, Paired Module means we adopt the 1:1 convolution and ReLU ratio in DFN-MR1 model.
Type1 Proportional Module means we remove the ReLU after element-wise in one branch,
Type 2 Proportional Module means we remove the ReLU before element wise in one branch.
We find that the Type 1 Proportional Module achieves the better performance on CIFAR-10 and CIFAR-100,
but the improvement is not higher enough. Type 2 Proportional Module cannot improve the performance compared with the baseline
\begin{table}[t]

	\centering
	\caption{Comparison between paired module and proportional module in DFN-MR1 on CIFAR-10/100.}
	\label{comparison}
	\begin{tabular}{|c|c|c|}
		\hline
		Model & Depth & Accuracy\\
        \hline
        \multicolumn{3}{|c|}{CIFAR-100} \\
		\hline
		{Paired Module \cite{Ref:Zhao2016}}  & {110} &  {75.59 $\pm$ 0.19} \\
		\hline
        {Type1 Proportional Module} & {110}  &   {\textbf{75.61 $\pm$ 0.38}}\\
        \hline
        {Type2 Proportional Module}  & {110}  &  {74.53 $\pm$ 0.80} \\
		\hline
        \multicolumn{3}{|c|}{CIFAR-10} \\
        \hline
        {Paired Module \cite{Ref:Zhao2016}}  & {110} &  {95.04 $\pm$ 0.06} \\
        \hline
        {Type1 Proportional Module}  & {110}  &   {\textbf{95.26 $\pm$ 0.14}}\\
        \hline
	\end{tabular}
\label{DFNT}
\end{table}

\section{Discussion}
In previous research, we empirically find that removing some ReLU layers in CNN models can achieve better performance.
Thus, in our work, we investigate the more flexible proportional module to design the better network architectures.

\subsection{The redundancy of ReLUs in Plain Network}
First, we think the redundancy of ReLUs is one of important reasons why the accuracy of deeper network gets saturated. In plain network, we find that removing the half of ReLUs
can great improve the classification performance. It verifies that the N:M (N$>$M) convolution and ReLU ratio is indeed effective, especially in deeper networks.
Therefore, 2:1 convolution and ReLU ratio may be a more suitable proportional module, but it is a question whether the module can improve the performance of other more popular networks
like ResNet.

\subsection{Comparison with ResNet and DFN Baseline}
Second, to validate the generalization of our method, we utilize ResNet and DFN as baseline to explore the influence of the proportional module. From our experimental results,
we find that the proportional module can help the building block, pre-activation bottleneck block and merge and run block achieve the better performance. Experimental
results show that the proportional module not only alleviates the degradation problem in plain network but also can further improve the performance of ResNet and DFN. It demonstrates
the effectiveness of proportional module in regularizing CNNs.

\subsection{To Further Explore the Influence of ReLU}
Third, from coarse to fine, we explore the influence of ReLUs in different positions on some blocks. For pre-activation bottleneck block, we find some interesting phenomena:
With the depth increasing, the influence of single ReLU in one block is reduced.
On the other hand, the influence of last ReLU in a pre-activation bottleneck block is the most important.
For the first phenomenon, we think that with the depth increasing, the redundancy of ReLU becomes serious so that the influence of single ReLU is reduced.
For the second phenomenon, we think that there are two reasons:
The one reason is that the first ReLU leads to information loss, which inhibits the learning effect of the first convolutional layer.
The other one is that two convolutional layers before the last ReLU can learn a better transformation of data. When this transformation of data was through ReLU,
the whole network can acquire the better non-linear information. In addition, from
our experimental results shown on Table \ref{DFNT}, we can find that the first ReLU in the merge and run block is more important than the ReLU after element-wise.
We think the first ReLU has an important effect on the representation ability of the block because the first ReLU controls the non-linearity of the whole block.

At last, we think that the reason why the proposed method works is two-fold. First, the redundancy of ReLUs leads to the accuracy of deeper network gets saturated.
Second, the proportional module utilizes information more effectively since it can recover the part of information discarded by ReLU.

\section{Conclusion}
In this work, we argue that the conventional paired module may not be the best choice to design the network architectures.
We found that proportional module can help network achieve better performance.
From our experimental results, we found that proportional module could enrich the expressive power of the network.
Although the non-linearity preserves the depth property of a network, the linear information cannot still be ignored for learning
a better distribution of data since it use information effectively. Experimental results show that such a simple design has superior
performance than the corresponding baselines while having the same computation cost compared to the original models. To our best knowledge, this
is the first work to introduce the proportional module in CNN models. For future work, we will utilize reinforcement learning to assemble adaptively the convolutional layer and ReLU.

\section*{Acknowledgments}
This work was supported in part by the National Key R\&D Program of China(No. 2018YFB1004600),
the National Natural Science Foundation of China (No. 61773375, No. 61375036, No. 61602481, No. 61702510),
and in part by the Microsoft Collaborative Research Project.
\bibliographystyle{IEEEtran}
\bibliography{IEEEabrv,ICPR}
\end{document}